\documentclass[letterpaper, 10 pt, journal, twoside]{IEEEtran} 
\IEEEoverridecommandlockouts

\usepackage{amsfonts}
\usepackage{mathtools}
\usepackage{diagbox}
\usepackage{amsmath}
\usepackage{color}
\usepackage{amssymb}
\usepackage{framed}
\usepackage{balance}
\usepackage{subfigure}
\usepackage{graphics} 
\usepackage{lipsum}
\usepackage{graphicx}
\usepackage{mathptmx} 
\usepackage{fancyhdr}
\usepackage[nospace]{cite}
\usepackage[ruled,vlined]{algorithm2e}
\usepackage{setspace}
\usepackage{amssymb}
\usepackage[table]{xcolor}
\usepackage{algpseudocode}
\usepackage{booktabs}
\usepackage{tabularx}
\usepackage{booktabs}
\usepackage{array}
\usepackage{etex}
\usepackage{adjustbox}
\usepackage{adjustbox,lipsum}
\usepackage{gensymb}
\usepackage{nicefrac}
\usepackage{mdwtab}
\usepackage{multirow}
\usepackage{colortbl}
\usepackage{hyperref}
\usepackage{color,soul}
\usepackage{epsfig}
\usepackage{tikz}
\usepackage{textcomp}
\usepackage{hyperref}
\usepackage{lipsum}
\DeclareMathAlphabet{\mathcal}{OMS}{cmsy}{m}{n}

\usepackage{stmaryrd}

\usepackage{etoolbox}
\makeatletter
\patchcmd{\@makecaption}
  {\scshape}
  {}
  {}
  {}
\makeatletter
\patchcmd{\@makecaption}
  {\\}
  {.\ }
  {}
  {}
\makeatother
\def\tablename{Table}

\algrenewcommand\algorithmicforall{\textbf{foreach}}
\algrenewcommand\algorithmicindent{.8em}

\usepackage{accents}

\usepackage{bm}

\DeclareMathOperator*{\argmax}{arg\,max}

\newcommand\copyrighttext{%
  \centering
  \footnotesize \textcopyright 2021 IEEE. Personal use is permitted, but republication/redistribution requires IEEE permission.
  DOI: \href{https://ieeexplore.ieee.org/document/9345363}{10.1109/LRA.2021.3056367}}
\newcommand\copyrightnotice{%
\begin{tikzpicture}[remember picture,overlay]
\node[anchor=south,yshift=10pt] at (current page.south) {\fbox{\parbox{\dimexpr\textwidth-\fboxsep-\fboxrule\relax}{\copyrighttext}}};
\end{tikzpicture}%
}

\begin{document}

\title{Task-Adaptive Robot Learning from Demonstration with Gaussian Process Models under Replication}

\author{Miguel Arduengo$^{1}$, Adri\`a Colom\'e$^{1}$, J\'ulia Borr\`as$^{1}$, Luis Sentis$^{2}$ and Carme Torras$^{1}$
\thanks{This letter was recommended for publication by Associate Editor S. Chernova and Editor D. Kulic upon evaluation of the reviewers’ comments. This work has been partially funded by the European Union Horizon 2020 Programme under grant agreement no. 741930 (CLOTHILDE) and by the Spanish State Research Agency through the María de Maeztu Seal of Excellence to IRI [MDM-2016-0656]..}
\thanks{$^{1}$First Author, Second Author, Third Author and Fifth Author are with Institut de Robòtica i Informàtica Industrial, CSIC-UPC (IRI), Barcelona. \{\tt\footnotesize marduengo, acolome, jborras, torras\}@iri.upc.edu}
\thanks{$^{2}$ Fourth Author is with the Human Centered Robotics Laboratory, University of Texas at Austin, Austin. {\tt\footnotesize lsentis@austin.utexas.edu}}}

\maketitle
\copyrightnotice

\begin{abstract}

Learning from Demonstration (LfD) is a paradigm that allows robots to learn complex manipulation tasks that can not be easily scripted, but can be demonstrated by a human teacher. One of the challenges of LfD is to enable robots to acquire skills that can be adapted to different scenarios. In this paper, we propose to achieve this by exploiting the variations in the demonstrations to retrieve an adaptive and robust policy, using Gaussian Process (GP) models. Adaptability is enhanced by incorporating task parameters into the model, which encode different specifications within the same task. With our formulation, these parameters can be either real, integer, or categorical. Furthermore, we propose a GP design that exploits the structure of replications, i.e., repeated demonstrations with identical conditions within data. Our method significantly reduces the computational cost of model fitting in complex tasks, where replications are essential to obtain a robust model. We illustrate our approach through several experiments on a handwritten letter demonstration dataset.   

\end{abstract} 


\section{INTRODUCTION}

\begin{figure}[t]
    \centering
    {\includegraphics[width=1.0\linewidth]{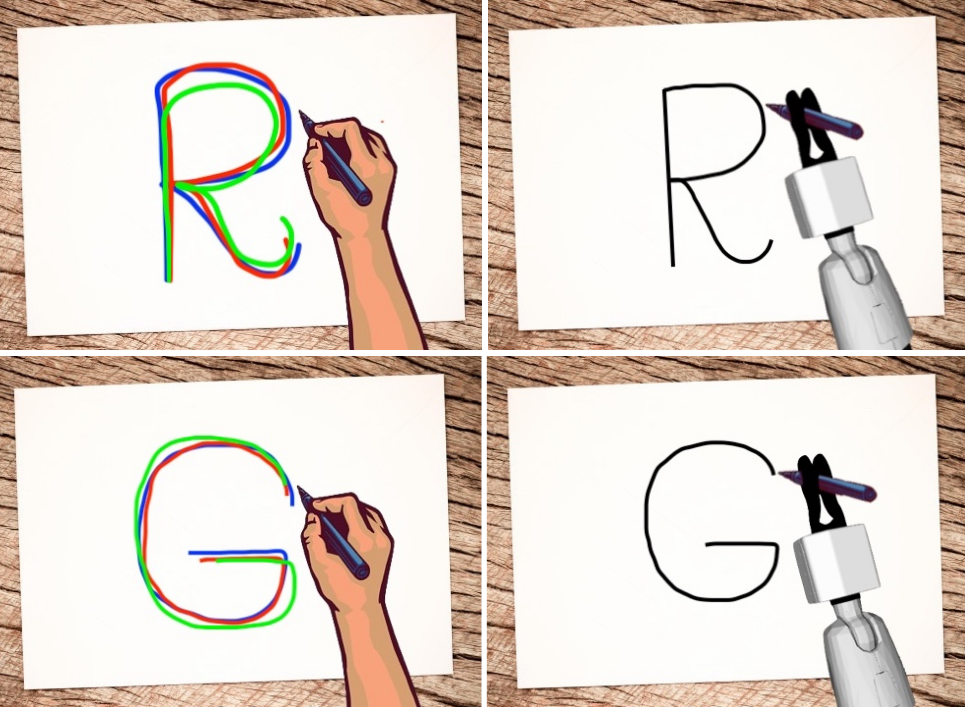}}
    \caption{Learning from Demonstration allows transferring the robot skills that can be intuitively demonstrated but are difficult to script, such as handwriting. On the left, the demonstrations. On the right, the learned policy.}
    \label{fig1}
    \vspace{-4mm}
\end{figure}

\IEEEPARstart{L}{earning} from Demonstration (LfD) is the paradigm in which robots implicitly learn task constraints from demonstrations. This allows more intuitive skill transfer, satisfying a need of opening policy development to non-robotic-experts as robots extend to assistive domains. The choice of LfD is particularly compelling when ideal behavior can be neither scripted nor easily defined as a reward function but can be demonstrated (Figure \ref{fig1}). One of the fundamental questions is \textit{What to imitate?} \cite{Nehaniv2011}. Trajectory-learning methods are usually adopted since they allow a direct skill transfer to robot actions, at both joint, and task space levels. Learning a manipulation task at a trajectory level involves modeling the set of demonstrated motions and retrieving a generalized representation. Among the most relevant contributions in the field over the past decade, we can highlight the approaches based on Dynamic Movement Primitives (DMP) \cite{Pastor2009}, Probabilistic Movement Primitives (ProMP) \cite{Paraschos2018}, Gaussian Mixture Models (GMM) \cite{Calinon2016}, Kernelized Movement Primitives (KMP)\cite{Huang2019b} and Gaussian Process models (GP) \cite{Delgado2020}. In a recent work \cite{Arduengo2020}, we presented a GP-based LfD framework, which we adopt as a basis for this paper. For a comparison with the aforementioned approaches, the reader can refer to our work.

The focus of this paper is on the generalization performance of the learned policy at a task level with GP models. There are skills, in which multiple demonstrations of the same task can look very different due to the task-specific variations. This variability can be interpreted to be governed by the so-called \textit{task variables}, which can describe the current context or a particular requirement. In the LfD literature, generalization has been mainly achieved with two distinct approaches: (a) encoding the demonstrations from the perspective of multiple reference frames; (b) considering task variables as inputs to the learned movement policy, enabling the generation of a path adapted to the context. Approach (a) is motivated by the observation that skillful movement planning often requires the orchestration of multiple coordinate systems that can have varying levels of importance along the task. In \cite{Calinon2016}, the authors propose Task-Parametrized GMM (TP-GMM), a direct extension of GMM. By providing a set of candidate frames that can be potentially relevant, a local statistical analysis is conducted to learn how to retrieve a general trajectory that is invariant to translations and rotations. The idea presented in TP-GMM is also applied to KMP in \cite{Huang2019a}. Using local coordinate systems, the robot is able to learn about the superposition and transition between the reference frames, resulting in improved extrapolation capabilities. This generalization performance comes at the expense of limiting the task parameters to be in the form of coordinate systems or local projection operators. On the other hand, approach (b) has been applied to DMP, ProMP, and GP. In \cite{Matsubara2010}, the authors present stylistic DMP (SDMP), which allows a compact encoding of diverse styles in the demonstrations by including a real-valued control variable in the model, called style parameter. Regarding ProMP, in \cite{Paraschos2018}, the authors propose to adapt the primitive based on an external real-valued state variable by learning a linear mapping to the mean weight vector. Finally, in \cite{Forte2012}, the authors model a map, from a target in 3-D space, to the reaching trajectory, with demonstrations using a GP model. Although (b) provides good interpolation results, the performance far from the regions covered by the demonstrations is limited. However, this approach is generic since the task variables can represent arbitrary context features.

Additionally, one should keep in mind that LfD is a supervised learning approach. Although extrapolation capabilities can be enhanced, providing enough demonstrations to cover the action space is essential. When the model is probabilistic (ProMP, GMM, KMP and GP), repetitions are also required for adequately inferring the statistics of the taught motion. That is, the demonstration dataset must sample the task variable space as densely as possible, providing as many replications as feasible. Thus, for modeling complex manipulation tasks the amount of required data might increase considerably. Commonly, LfD methods are applied with a dataset of a few demonstrations, since they do not scale very well. On the one hand, ProMP and GMM involve an Expectation-Maximization algorithm. On the other hand, KMP and GP require a matrix inversion, whose dimension increases with the number of data points. In this paper, we focus on movement policies learned by demonstration using GP. The approaches for improving the scalability of GP while retaining favorable prediction quality can be divided into two groups: sparse (or global) approximations \cite{Snelson2005}, where the training dataset is approximated by a smaller set of so called support points; and local approximations \cite{Schneider2010a}, where the dataset is divided into subsets, using only points near the desired input location for making the prediction. The main drawback of these solutions is their approximate nature.

The main contributions of the paper are two-fold: (1) a GP design for including task variables in the model, which can be either real, integer or categorical; (2) the exploitation of the structure of replications, for alleviating the computational complexity of GP while retrieving an exact model. Our aim is to extend our previous work on LfD with GP by enhancing its generalization to variant task conditions and its capability for scaling with a large demonstration dataset. The structure of the paper is organized as follows: in Section \ref{problem-statement} we formally state the considered LfD problem; in Section \ref{GP-based LfD} we outline the main theoretical aspects of GP-based LfD; then in Section \ref{task-adaptive-under-rep} we present our approach for including task variables and exploiting replications; next, in Section \ref{experiment}, we illustrate the main concepts and analyze the performance of the proposed solution by learning the writing task from a handwritten letter dataset; finally, in Section \ref{conclusion} we summarize the main conclusions.

\section{LfD Problem Statement}
\label{problem-statement}

We formally construct the LfD problem as follows. The robot is presented with a demonstration dataset:
\begin{equation}
    \mathcal{D}=\left\{\bm{x}_{ij},\,\bm{y}_{ij}\right\}_{i,\;j=1}^{N,\;M_i}
\end{equation} where $\bm{x}_{ij}\in\mathcal{X}$ denotes the task variables and $\bm{y}_{ij}\in\mathbb{R}^{\mathcal{O}}$ the variables describing the demonstrated motion. Here, the super-indexes $N$, $M_i$ and $\mathcal{O}$, correspond respectively to the number of demonstrations, training samples per demonstration (which can vary) and the output dimension. Assuming that one of the input components is time $t$,  $\mathcal{X}=\mathbb{R}\times\prod_{i=2}^\mathcal{I}\mathbb{X}_i$, being $\mathcal{I}$ the input dimension and $\mathbb{X}_i$ the subspace where the $i$-th component lies. The LfD problem is solved by learning a task movement policy  $\pi:\mathcal{X}\longrightarrow\mathbb{R}^{\mathcal{O}}$ from $\mathcal{D}$, which is capable of inferring the required path to successfully perform the desired manipulation given a new set of task variables. The policy must generalize over multiple demonstrations, and also, in order to be reproduced by the robot, the generated paths have to be continuous and smooth. 

\section{Gaussian-Process-based LfD}
\label{GP-based LfD}

\begin{figure*}[t]
    \centering
    {\includegraphics[width=1.0\linewidth]{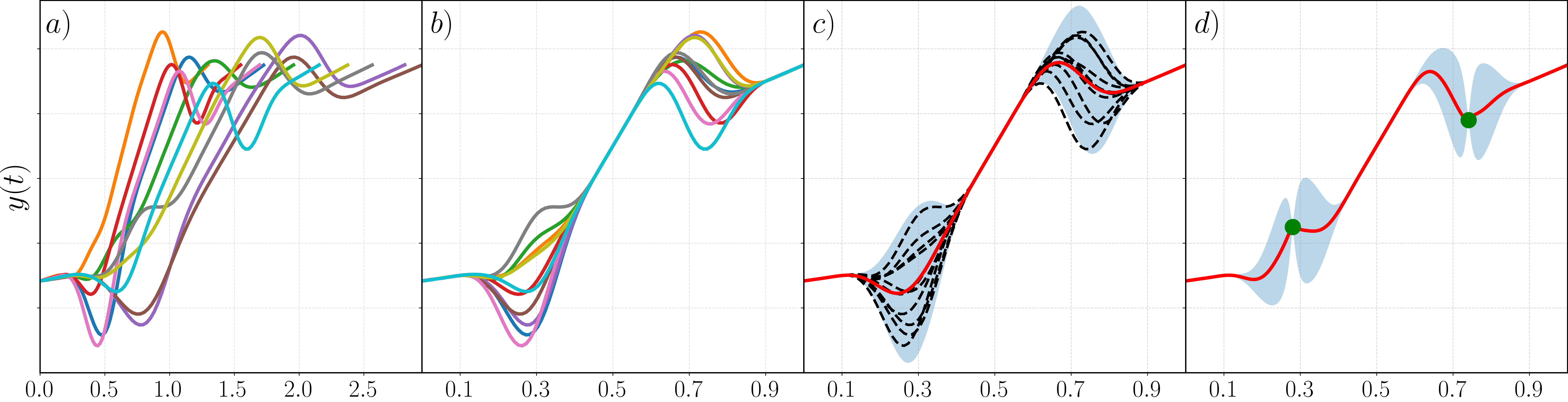}}
    \caption{Learning from demonstration with Gaussian Processes. \textbf{(a)} Several demonstrations of the manipulation task. The trajectories present spatial variability as well as different time lengths. \textbf{(b)} They are temporally aligned with the DTW algorithm using the TCI as similarity measure. Also, the time component is scaled, so the trajectory is adapted to the desired time law. \textbf{(c)} The movement policy is encoded with a Heteroscedastic GP model. This probabilistic representation effectively encodes the variability (blue shaded area) of the demonstrations (black dashed line) and retrieves a generalized form of the task motion (red solid line). \textbf{(d)} The learned policy can be modulated through via-points (green dots) by conditioning the policy on the new points.}
    \label{fig2}
    \vspace{-5mm}
\end{figure*}

Gaussian Process are a probabilistic representation that allow to encode the underlying trajectory distribution from multiple demonstrations of a manipulation task. For simplicity, in this section we propose to use $\mathcal{D}=\left\{t_{ij},\,\bm{y}_{ij}\right\}_{i,\;j=1}^{N,\;M_i}$ i.e. the only input component is time. A high-level description of the GP-based LfD framework is summarized in Figure \ref{fig2}. We first perform a preprocessing step for temporally aligning and scaling the demonstrated trajectories. For the training phase, we provide a series of GP design guidelines for modeling robot manipulation tasks from $\mathcal{D}$. Finally, we briefly discuss how the learned policy can be modulated through via-points during the movement execution phase.

\subsection{Preprocessing: Temporal Alignment and Scaling}

In general, it is difficult to provide all the demonstrations with the same speed. Time shifts might lead to poor performance when inferring the variability. For aligning temporally the demonstrated trajectories Dynamic Time Warping (DTW) \cite{Senin2008} can be used. It is a method that calculates an optimal match, usually nonlinear, with respect to a reference trajectory based on a similarity measure. We use the Task Completion Index (TCI) \cite{Arduengo2020}, that is a measure of the portion of the trajectory covered for task completion. For demonstration $i$, it can be computed as
\begin{equation}
    \zeta_{i,k} = \frac{\sum_{j=1}^k\lVert \bm{y}_{i,j}-\bm{y}_{i,j-1}\rVert}{\sum_{j=1}^{M_i}\lVert\bm{y}_{i,j}-\bm{y}_{i,j-1} \rVert}\quad\quad \forall k=1,\dots,M_i
\end{equation} 

Adjusting the execution speed of the robot is sometimes desirable. As for DMP and ProMP, we can consider a normalized phase variable to decouple the path from the time signal. This is equivalent to scale the time component. Let $t_R$ be the time length of the reference trajectory. To adapt the trajectory to a desired duration $t_D$, we can define a monotonic increasing function $s:\left[0,\,t_R\right]\longrightarrow\left[0,\,t_D\right]$. Note that this step can also be carried out during the execution.

\subsection{Robot Task Representation with GP}

The task policy must encode the variability as well as generalize over multiple demonstrations of the learned manipulation task. This requires an adequate GP design.  

\subsubsection{Gaussian Process Models} Intuitively, a GP defines a prior over functions, which can be converted into a posterior given a set of input-output pairs \cite{Rasmussen2006}. Considering scalar outputs $y$, it is defined by the scalar mean $m\left(t\right)$ and covariance $k\left(t,t'\right)$ (i.e. kernel) functions, which encode the assumptions on the policy being learned
\begin{equation}
    y\left(t\right)\sim \mathcal{GP}\left(m\left(t\right),\,k\left(t,t'\right)\right)
\end{equation} 

\subsubsection{Heteroscedastic GP} The task policy i.e. the path $\bm{y}^*$ that performs the taught manipulation given a new set of task variables $\bm{t}^*$ and $\mathcal{D}$, is modeled with GP as a multivariate Gaussian distribution $p\left(\bm{y}^*|\bm{t}^*,\mathcal{D}\right)\sim \mathcal{N}\left(\boldsymbol{\mu}^*,\boldsymbol{\Sigma}^*\right)$. The constant noise level considered in the standard GP formulation can be an important limitation for capturing the variability in the demonstrations, as there will be parts in the task where it might vary. The uniform noise assumption can be relaxed by considering a normally distributed noise $\epsilon\sim\mathcal{N}\left(0,r\left(t\right)\right)$, where the variance is input-dependent and modelled by the latent function $r()$. The joint distribution of the training $\bm{y}$, and predicted $\bm{y}^*$ outputs according to the prior is then
\begin{equation}
\small
\label{joint-eq}
    \left[\begin{array}{c}
         \bm{y}  \\
         \bm{y}^* 
    \end{array}\right]\sim \mathcal{N}\left(\left[\begin{array}{c}
         \bm{m}  \\
         \bm{m}^*
    \end{array}\right],\left[\begin{array}{cc}
         \bm{K}+\bm{R} & \bm{K}^*\\
         \bm{K}^{*T} & \bm{K}^{**}+\bm{R}^*
    \end{array}\right]\right)
\end{equation}\normalsize being $\bm{m}$ and $\bm{m}^*$ arrays whose elements are function $m()$ evaluated at $\bm{t}$ and $\bm{t}^*$ respectively; analogously for diagonal matrices $\bm{R}$ and $\bm{R}^*$ with function $r()$; $\bm{K}^*$ is the Gram matrix of the kernel function $k(,)$ evaluated at all pairs $\left(t,t^*\right)$; similarly, $\bm{K}$ and $\bm{K}^{**}$. The predictive distribution is then obtained by marginalizing on the demonstrations, resulting the following predictive mean $\boldsymbol{\mu}^*$ and variance $\boldsymbol{\Sigma}^*$
\begin{align}
    &\boldsymbol{\mu}^*=\bm{m}^*+\bm{K}^{*T}\left(\bm{K}+\bm{R}\right)^{-1}\left(\bm{y}-\bm{m}\right) \label{mean-pred}\\
    &\boldsymbol{\Sigma}^*=\bm{K}^{**}+\bm{R}^*-\bm{K}^{*T}\left(\bm{K}+\bm{R}\right)^{-1}\bm{K}^{*}\label{cov-pred}
\end{align}

The latent noise function $r()$ is usually not known a-priori. As proposed in \cite{Snelson2006}, first an standard GP can be fit to the demonstrations. Its predictions can be then used to estimate the input-dependent noise empirically. Then, a second GP can  be used to model $z\left(t\right) =\log\left[r\left(t\right)\right]$. Let $\mathcal{Z}=\left\{\bm{z},\bm{z}^*\right\}$ be the set of noise data. The posterior predictive distribution can be then approximated by
\begin{equation}
p\left(\bm{y}^*|\bm{t}^*,\mathcal{D}\right)\simeq p\left(\bm{y}^*|\bm{t}^*,\mathcal{D},\mathcal{Z}\right)\end{equation} where $\mathcal{Z}=\argmax_{\mathbf{z},\mathbf{z}^*}p\left(\bm{z},\bm{z}^*|\bm{t}^*,\mathcal{D}\right)$ is the most likely noise level, which can be determined with Monte Carlo or Expectation-Maximization algorithms.

\subsubsection{Multi-Output GP (MOGP)} The previous concepts can be extended to multiple-output GP (MOGP) by taking a matrix covariance function $\bm{k}\left(t,t'\right)$. This can be expressed around the Linear Model of Coregionalization (LMC) \cite{Alvarez2012}
\begin{equation}
\small
    \bm{B}\otimes\bm{k}\left(t,t'\right)=\left[\begin{array}{ccc}
         B_{11}k_{11}\left(t,t'\right) & \dots & B_{1d}k_{1d}\left(t,t'\right) \\
         \vdots & \ddots & \vdots \\
         B_{d1}k_{d1}\left(t,t'\right) & \dots & B_{dd}k_{dd}\left(t,t'\right)
    \end{array}\right]
\end{equation}\normalsize where $d$ is the output dimension and $\bm{B}$ is the coregionalization matrix. The off-diagonal elements encode output relatedness. In the general case, for learning robot task motions, we cannot make any a-priori assumptions in this regard. Therefore, we can set $B_{ij}=0$ for $i\neq j$, being the MOGP equivalent to $d$ independent GP.

\subsubsection{Kernel}
In order to be a valid kernel function, the corresponding Gram matrix $\boldsymbol{K}$, with elements $K_{ij}=k\left(t_i,\,t_j\right)$ must be positive semidefinite. Furthermore, the chosen kernel must generate continuous and smooth paths for the robot to be able to execute the motion. Note also that the time parametrization of trajectories is invariant to translations in the time domain. Thus, it should be a function of $\tau=\lVert t-t'\rVert$. A popular choice is the squared exponential (SE) kernel
\begin{equation}
\label{squared_exponential}
\small
    k\left(\tau\right)=\sigma_f^2\text{exp}\left(-\frac{\tau^2}{2l^2}\right)
\end{equation} where $\sigma_f$ and $l$ are the hyperparameters. The GP with this covariance function has mean square derivatives of all orders, and is thus very smooth. For slightly relaxing the smoothness prior assumption, the Mat\'ern kernel can also be used
\begin{equation}
\small
\label{matern}
    k\left(\tau\right)=\sigma_f^2\left(1+\frac{\sqrt{5}\tau}{l}+\frac{5\tau^2}{3l^2}\right)\text{exp}\left(-\frac{\sqrt{5}\tau}{l}\right)
\end{equation}

For selecting the kernel hyperparameters $\boldsymbol{\Theta}$, the following log marginal likelihood is usually maximized
\begin{equation}
\label{likelihood}
    \log p\left(\bm{y}|\bm{t},\boldsymbol{\Theta}\right)=-\frac{1}{2}\bm{y}^T\left(\bm{K}+\bm{R}\right)^{-1}\bm{y}-\frac{1}{2}\log|\bm{K}+\bm{R}|
\end{equation} 

This optimization problem might suffer from multiple local optima. Gradient-based or 
stochastic optimization methods can be used for computing the solution.

\subsection{Policy Modulation with GP} 

We can modulate the task policy by adapting the learned path to pass trough new via-points  $\mathcal{V}=\left\{t_i,\bm{y}^{v}_i\right\}_{i=1}^{M_v}$. Modulation can be achieved by conditioning the policy on both, $\mathcal{D}$ and $\mathcal{V}$. Assuming that the predictive distribution of each set can be computed independently
\begin{equation}
    p\left(\bm{y}^*|\bm{t}^*,\mathcal{D},\mathcal{V}\right) \propto p\left(\bm{y}^*|\bm{t}^*,\mathcal{D}\right)p\left(\bm{y}^*|\bm{t}^*,\mathcal{V}\right)
\end{equation} 

Then, if $p\left(\bm{y}^*|\bm{t}^*,\mathcal{D}\right)\sim\mathcal{N}\left(\boldsymbol{\mu}^d,\boldsymbol{\Sigma}^d\right)$ and $p\left(\bm{y}^*|\bm{t}^*,\mathcal{V}\right)\sim\mathcal{N}\left(\boldsymbol{\mu}^v,\boldsymbol{\Sigma}^v\right)$, it holds $p\left(\bm{y}^*|\bm{t}^*,\mathcal{D},\mathcal{V}\right)\sim\mathcal{N}\left(\boldsymbol{\mu}^{**},\boldsymbol{\Sigma}^{**}\right)$ with 
\begin{align}
    &\boldsymbol{\mu}^{**}=\boldsymbol{\Sigma}^v\left(\boldsymbol{\Sigma}^d+\boldsymbol{\Sigma}^v\right)^{-1}\boldsymbol{\mu}^d+\boldsymbol{\Sigma}^d\left(\boldsymbol{\Sigma}^d+\boldsymbol{\Sigma}^v\right)^{-1}\boldsymbol{\mu}^v \\
    &\boldsymbol{\Sigma}^{**}=\boldsymbol{\Sigma}^d\left(\boldsymbol{\Sigma}^d+\boldsymbol{\Sigma}^v\right)^{-1}\boldsymbol{\Sigma}^v
\end{align} The resulting policy ponders the demonstrations and the via-points weighted inversely by their variances. Modelling $p\left(\bm{y}^*|\bm{t}^*,\mathcal{V}\right)$ with an heteroscedastic GP, the strength of the via-point constraints can be easily specified by means of $r()$. Note that modulation can be performed during the execution of the motion, since $p\left(\bm{y}^*|\bm{t}^*,\mathcal{D}\right)$ can be pre-computed.

\section{Task-Adaptive GP under Replication}
\label{task-adaptive-under-rep}

Our novel GP-based framework takes advantage of the versatility and expressiveness of GP to encompass the main features required for a state-of-the-art LfD approach \cite{Arduengo2020}. Generalization of the learned policy can be enhanced by incorporating task variables that describe the context under which demonstrations are performed. We propose a new GP design inspired by \cite{Roustant2020}, which allows these variables to be either real, integer or categorical, and exploits the possible correlations. Also, for complex tasks, the number of required demonstrations for sampling the action space might increase considerably. This poses a challenge for GP, which suffer from cubic complexity to data size. Motivated by \cite{Binois2017}, we propose a formulation that exploits the structure of replications, which arise naturally in the LfD context, for achieving significant computational savings. 

\subsection{Generalization with Task Variables}

Encoding the policy with GP, we can consider task variables as inputs. In this way, relying on $\mathcal{D}$, the model can learn the constraints and requirements of the manipulation task from a wider perspective, being capable of retrieving an adaptive motion for a previously unseen context, described by a new set of task variables. However, the standard GP problem considers only continuous input variables, limiting the applicability of the method in tasks with discrete integer or categorical variables (e.g. object class, house room).

As a first approach, we could fit distinct GP models for each possible combination of the discrete variables. However, this method ignores possible correlations and becomes infeasible as the quantity of discrete sets grows, since the number of models increases exponentially. Another possibility is one-hot encoding i.e. adding as many extra input variables as different values the discrete variable can take. Although it might be appealing for its simplicity, the input dimension can increase dramatically.

Without loss of generality, we consider a three-dimensional input variable $\bm{x}=\left(t,\,s,\,u\right)$, being $t$ continuous, $s$ integer and $u$ categorical variables. Thus, we study GP models defined on a finite subspace of $\mathcal{X}=\mathbb{R}\times\mathbb{Z}\times\mathbb{K}$. By focusing the modeling effort on the covariance structure, kernels on $\mathcal{X}$ can be obtained by combining kernels on $\mathbb{R}$, $\mathbb{Z}$ and $\mathbb{K}$. Standard valid combinations are the (1) product, (2) sum or (3) ANOVA. If $k_{\mathbb{X}}$ denotes a kernel for variables that lie in domain $\mathbb{X}$, examples of valid kernels are
\begin{enumerate}
    \item $k\left(\bm{x},\bm{x}'\right)=k_{\mathbb{R}}\left(t,t'\right)k_{\mathbb{Z}}\left(s,s'\right)k_{\mathbb{K}}\left(u,u'\right)$
    \item $k\left(\bm{x},\bm{x}'\right)=k_{\mathbb{R}}\left(t,t'\right)+k_{\mathbb{Z}}\left(s,s'\right)+k_{\mathbb{K}}\left(u,u'\right)$
    \item $k\left(\bm{x},\bm{x}'\right)=\left[1+k_{\mathbb{R}}\left(t,t'\right)\right]\left[1+k_{\mathbb{Z}}\left(s,s'\right)\right]\left[1+k_{\mathbb{K}}\left(u,u'\right)\right]$
\end{enumerate}

For $k_{\mathbb{R}}$, kernels such as the the squared exponential (Eq. \ref{squared_exponential}) or Mat\'ern (Eq. \ref{matern}) can be used. The question then comes down to constructing a valid kernel on a finite subset of $\mathbb{Z}$ or $\mathbb{K}$, with a corresponding positive semidefinite Gram matrix. 

\subsubsection{Kernels for integer variables} An integer variable is a discrete variable with ordered levels. Thus, $\mathbb{Z}$ can be seen as a discretization of $\mathbb{R}$. We can define a non-decreasing transformation $T:\mathbb{Z}\longrightarrow\mathbb{R}$ (i.e. warping) such that the order is preserved, which projects the discrete variable into a continuous space. Consequently, the kernel function can be written as 
\begin{equation}
    \label{eq-linear-warping}
    k_{\mathbb{Z}}\left(s,s'\right)=k_{\mathbb{R}}\left(T\left(s\right),\,T\left(s'\right)\right)
\end{equation}
 
In the general case, $T$ is piecewise-linear. However, common warping functions are based on the cumulative distribution of a uniform, normal or lognormal random variable $T:\mathbb{Z}\longrightarrow\left[0,\,1\right]$. Note that when $k_{\mathbb{R}}$ depends on the distance $\lVert t-t' \rVert$, then $k_{\mathbb{Z}}$ depends on the distance between $s$, $s'$ distorted by $T$. Selecting an appropriate $T$ may require subject-matter knowledge. 

In order to allow negative correlations, alternatively to standard SE or Mat\'ern kernels, one may choose, for instance, the cosine correlation kernel on $\left[0,\,\beta\right)$, being $\beta\in\left(0,\,\pi\right]$ a fixed parameter tuning the minimal correlation value
\begin{equation}
\label{cosine}
    k_{\mathbb{Z}}\left(s,\,s'\right)=\cos{\left(T\left(s\right)-T\left(s'\right)\right)}
\end{equation}

\subsubsection{Kernels for categorical variables} For categorical variables there is no notion of order. However, what does exist is a notion of equality ($=$) or inequality ($\neq$). Among the parsimonious kernel parametrizations, up to additional assumptions, which generate a positive-definite Gram matrix is the compound symmetry (CS) covariance structure 
\begin{equation}
\label{cs-standard}
    k_{\mathbb{K}}\left(u,u'\right)=\begin{cases}
        v \quad \text{if }u=u' \\
        c \quad \text{if }u\neq u' 
    \end{cases}%\quad c/v\in\left(-1/\left(L-1\right),1\right)
\end{equation} where $v$ is the variance and $c$ the covariance. This structure is a generalization of the SE kernel using the Gower distance. All pairs of categories are treated equally, being the similarity maximum for two equal input points, and minimum for different ones. A more flexible parametrization can be obtained by considering groups of categories. Let the discrete categorical set be partitioned into $G$ groups, and $g(u)$ the group number corresponding to value $u$
\begin{equation}
        k_{\mathbb{K}}\left(u,u'\right)=\begin{cases}
        v \qquad\quad\quad\, \text{if }u=u' \\
        c_{g(u),g(u')} \quad \text{if }u\neq u' 
    \end{cases}
\end{equation} where for all $i,j\in\left\{1,\dots,G\right\}$, the terms $c_{i,i}/v$ are within-group correlations, and $c_{i,j}/v\;\left(i\neq j\right)$ are between-group correlations. Note that additional constraints on $v$ and $c_{i,j}$ are required to ensure that $k_{\mathbb{K}}$ is a valid kernel function. The corresponding Gram matrix $\bm{K}$, written in block form is
\begin{equation}
    \bm{K}\left(\bm{u},\bm{u}'\right)=\left(\begin{array}{ccc}
         \bm{W}_1 & \hdots & \bm{B}_{1,G} \\
         \vdots & \ddots & \vdots \\
         \bm{B}_{G,1} & \hdots & \bm{W}_G
    \end{array}\right)
\end{equation} where diagonal blocks $\bm{W}_g$ and off-diagonal blocks $\bm{B}_{g,g'}$ encode within-group and  between-group covariances respectively. The necessary and sufficient conditions for $\bm{K}$ to be positive semidefinite are:
\begin{itemize}
    \item \makebox[2cm][l]{$\bm{W}_g$} is positive semidefinite $\forall\;g=1,\dots,G$ 
    \item \makebox[2cm][l]{$\bm{W}_g-\overline{W}_g\bm{J}_{g}$} is positive semidefinite $\forall\;g=1,\dots,G$ 
\end{itemize} where $\bm{J}_{g}$ is a matrix of ones with the size of $\bm{W}_g$ and $\overline{W}_g$ the average of its elements. Thus, for the kernel function defined in Equation (\ref{cs-standard}), considering a discrete set with $L$ different categories, we can derive the following condition
\begin{equation}
    -\left(L-1\right)^{-1}v<c<v
\end{equation}

\subsubsection{Example} 
\begin{figure}[h]
    \vspace{-2mm}
    \centering
    {\includegraphics[width=1.0\linewidth]{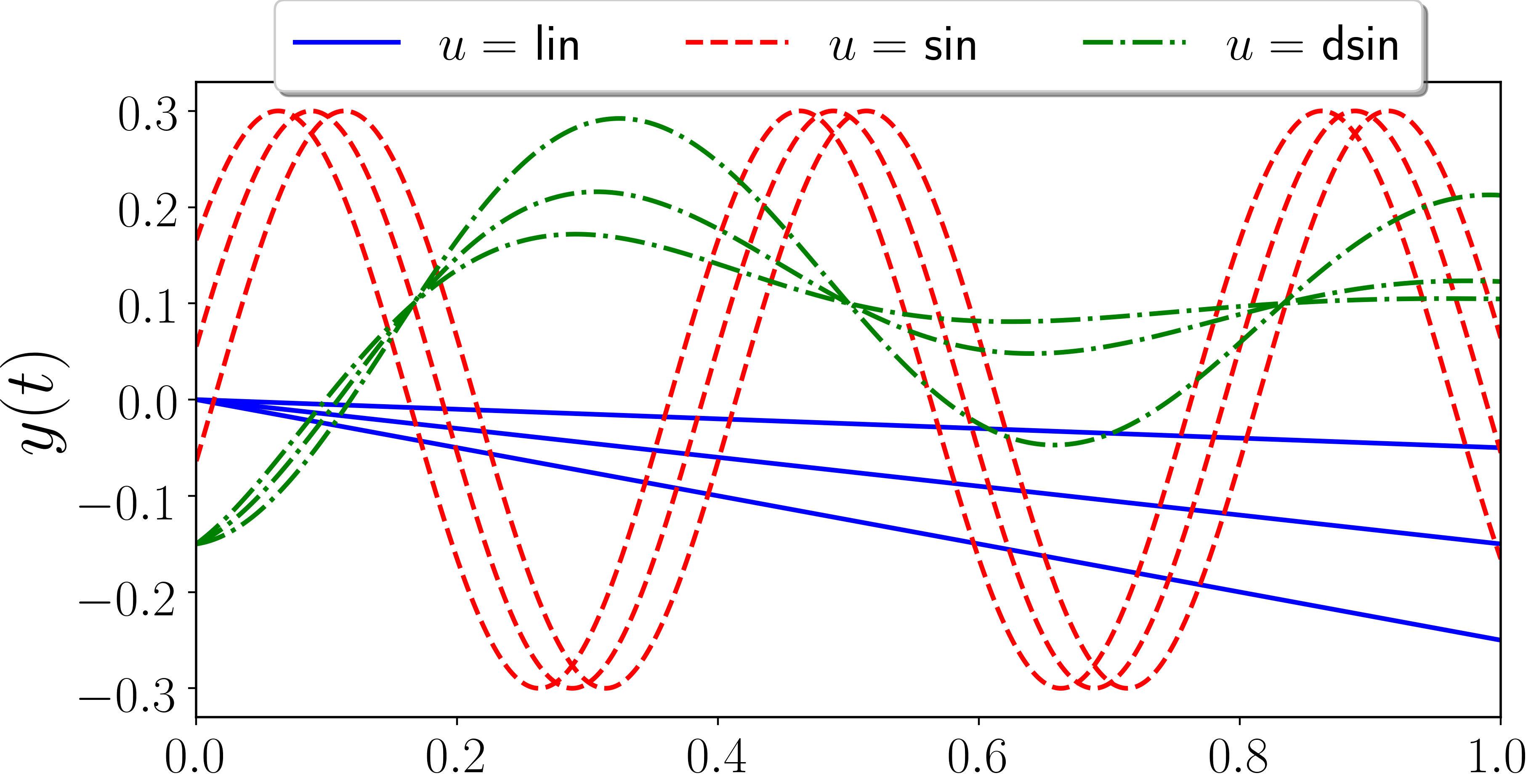}}
    \caption{Projection on the $y$-$t$ plane of the task model $y\left(\bm{x}\right)$ (Eq. \ref{example-fun})}
    \label{fig3}
\end{figure}

Consider a manipulation task modeled by the following deterministic function of $\bm{x}=\left(t,s,u\right)$ 
\begin{equation}
    \label{example-fun}
    y\left(\bm{x}\right)=\begin{cases}
    -\frac{1}{20}t\cdot s&\text{if } u=\text{lin}\\
    \frac{3}{10}\sin\left(\pi\left[5t-\frac{1}{4}\right]-s/5\right) &\text{if } u=\text{sin}\\
    \frac{1}{4}\sin\left(\pi\left[3t-\frac{1}{2}\right]\right)e^{-0.8t\cdot s}+\frac{1}{10} &\text{if } u=\text{dsin}
    \end{cases}
\end{equation} with $t\in\left[0,1\right]$, $s\in\left\{1,\dots,5\right\}$ and $u\in\left\{\text{lin},\text{sin},\text{dsin}\right\}$. As depicted in Figure \ref{fig3}, the required path depends greatly on $u$, while small variations are explained by $s$.

\begin{figure}[t]
    \begin{minipage}[b]{0.65\linewidth}
    \vspace{0pt}
    \centering
    \includegraphics[width=0.9\linewidth]{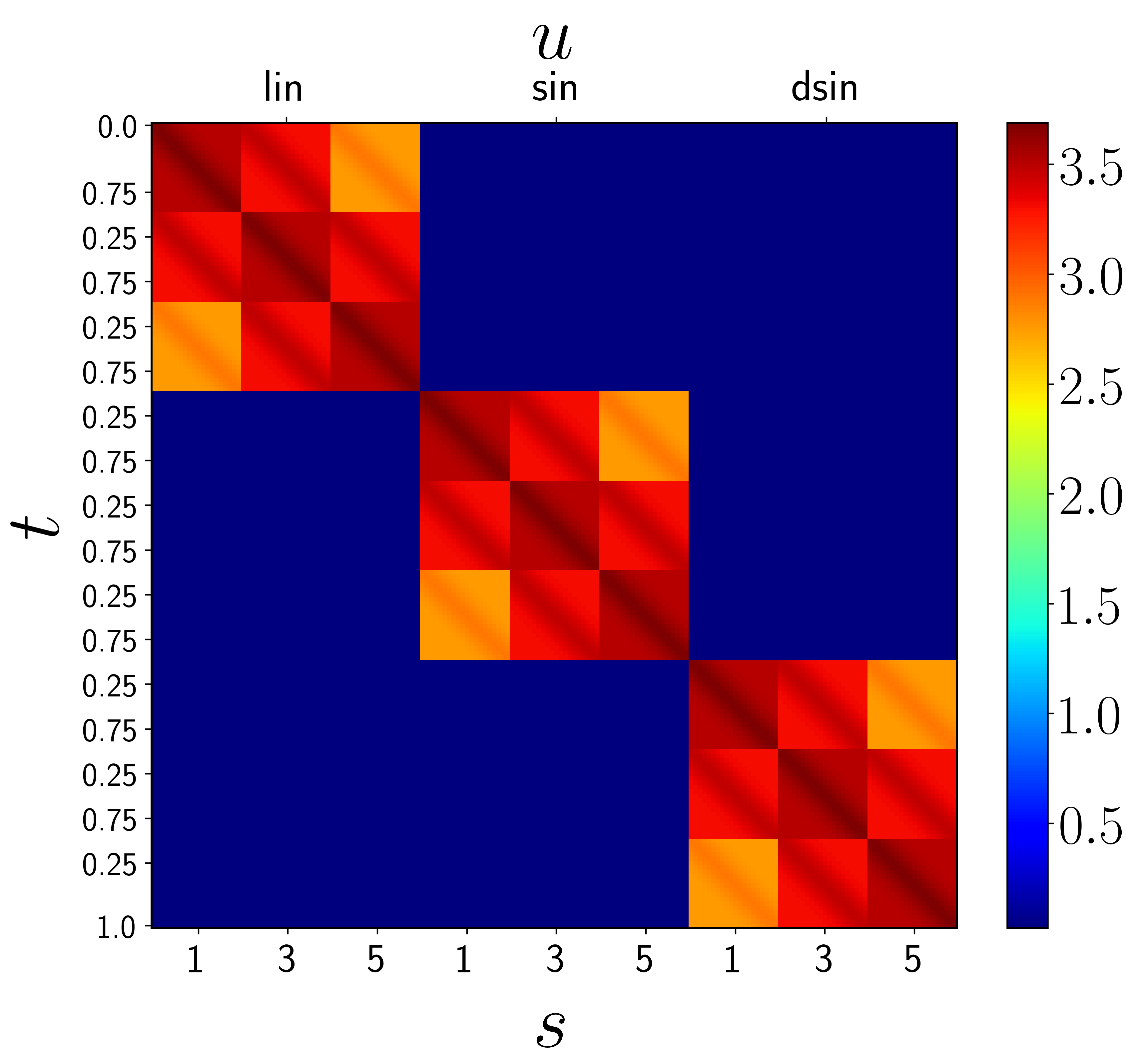}
    \end{minipage}
    \;
    \begin{minipage}[b]{0.25\linewidth}
    \vspace{0pt}
    \centering
    \begin{tabular}[b]{||c|c||}\hline
            \footnotesize{\textbf{Kernel}} & \footnotesize{$\boldsymbol{R^2}$} \\ \hline
            \footnotesize{product} & \footnotesize{0.52} \\
            \footnotesize{sum} & \footnotesize{0.37} \\
            \footnotesize{ANOVA} & \footnotesize{0.97} \\ \hline
        \end{tabular}
        %\label{tabr2}
    \vspace{13mm}
    \end{minipage}
    \vspace{-4mm}
    \caption{Matrix $\bm{K}$ constructed with ANOVA composition and maximum likelihood over a train set formed by a regular $8\times 3\times 3$ on $\left(t,s,u\right)$ domain. Also, $R^2$ coefficient of the inferred $y(\bm{x})$ using different compositions.}
    \label{fig4}
    \vspace{-6mm}
\end{figure}

We aim at reconstructing the task policy from demonstrations. As training data, we take samples on a regular $8\times3\times3$ grid in the task variable domain $\left[0,1\right]\times\left\{1,\dots,5\right\},\times\left\{\text{lin},\text{sin},\text{dsin}\right\}$. We consider three models using sum, product and ANOVA kernel combinations. For $k_{\mathbb{R}}$, $k_{\mathbb{Z}}$ and $k_{\mathbb{K}}$ we use the SE (Eq. \ref{squared_exponential}), cosine (Eq. \ref{cosine}) and CS (Eq. \ref{cs-standard}) kernels respectively. Hyperparameters are estimated by maximum likelihood. Model accuracy is measured in terms of the $R^2$ criterion over a test set formed by a finer $100\times 5\times 3$ grid in order to illustrate the generalization capabilities. In Figure \ref{fig4} we show the resulting matrix $\bm{K}$ with ANOVA composition, for which we achieve the best score. We can observe three distinct groups almost uncorrelated for different values of $u$, each one divided into three subgroups, one for each training $s$, with high relatedness, as it can be intuited from Figure \ref{fig3}. Thus, we can conclude that the proposed GP design retrives an effective policy by including continous, integer and categorical variables; inferring an accurate covariance structure.

\subsection{Inference and Prediction under Replication}

In the LfD context, replications, which can be intuitively defined as repeated demonstrations for identical task variables, play a key role on the estimation of the variability. This constitutes a challenge, since the computational complexity of the GP-based policy learning algorithm increases cubically with the number of training samples. We exploit the structure of replications in GP design for achieving computational savings during inference and prediction by using two well-known formulas, together comprising the Woodbury identity
\begin{align}
    &\left(\bm{D}+\bm{U}\bm{B}\bm{V}\right)^{-1}=\bm{D}^{-1}-\bm{D}^{-1}\bm{U}\left(\bm{B}^{-1}+\bm{V}\bm{D}^{-1}\bm{U}\right)^{-1}\bm{V}\\
    &\left|\bm{D}+\bm{U}\bm{B}\bm{V}\right|=\left|\bm{B}^{-1}+\bm{V}\bm{D}^{-1}\bm{U}\right|\cdot\left|\bm{B}\right|\cdot\left|\bm{D}\right|
\end{align}

Let $N$ and $n\ll N$ be the number of training samples and unique input locations (points with identical task variables) respectively; $y_i^{(j)}$ be the $j^{\text{th}}$ out of $a_i\geq 1$ replicates observed at each unique input, where $\sum_{i=1}^n a_i=N$; and $\overline{\bm{y}}$ be the array with concatenated terms $\overline{y}_i=a_i^{-1}\sum_{j=1}^{a_i}y_i^{(j)}$. We now develop a map from full $\bm{K}_{N}$, $\bm{R}_{N}$ matrices (Eq. \ref{joint-eq}) to their unique-$n$ counterparts $\bm{K}_{n}$, $\bm{R}_{n}$. Without loss of generality, assume that data is ordered such that replicates are stacked together. Then
\begin{equation}
    \bm{K}_{N}=\bm{U}\bm{K}_n\bm{U}^T\qquad\quad \bm{U}^T\bm{R}_{N}\bm{U}=\bm{A}_n\bm{R}_n
\end{equation} with $\bm{U}=\text{diag}\left(\bm{1}_{a_1,1}\dots,\bm{1}_{a_n,1}\right)$ a $N\times n$ block matrix, where $\bm{1}_{k,l}$ is a $k\times l$ matrix of ones, and $\bm{A}_n=\text{diag}\left(a_1,\dots,a_n\right)$. Developing equations \ref{mean-pred}, \ref{cov-pred} and \ref{likelihood}, taking $\bm{D}\equiv\bm{R}_{N}$, $\bm{B}\equiv\bm{K}_{n}$ and $\bm{V}=\bm{U}^T$, yields the following predictive equations
\begin{align}
    &\boldsymbol{\mu}^*=\bm{m}^*+\bm{K}_n^{*T}\left(\bm{K}_n+\bm{A}_n^{-1}\bm{R}_n\right)^{-1}\left(\overline{\bm{y}}-\overline{\bm{m}}\right) \\
    &\boldsymbol{\Sigma}^*=\bm{K}^{**}+\bm{R}^*-\bm{K}_n^{*T}\left(\bm{K}_n+\bm{A}_n^{-1}\bm{R}_n\right)^{-1}\bm{K}_n^{*}
\end{align} Note they are almost identical to the original ones built by overlooking the structure of replication. We also have the next expression for the log likelihood $\log\mathcal{L}$
\begin{multline}
    \log\mathcal{L}=-\frac{1}{2}\left(\bm{y}^T\bm{R}_{N}^{-1}\bm{y}-\overline{\bm{y}}^T\bm{A}_n\bm{R}_n^{-1}\overline{\bm{y}}+\overline{\bm{y}}^T\left(\bm{K}_n+\bm{A}_n^{-1}\bm{R}_n\right)^{-1}\overline{\bm{y}}\right) \\
    -\frac{1}{2}\left(\log\left|\bm{K}_n+\bm{A}_n^{-1}\bm{R}_n\right|+\log\left|\bm{R}_{N}\right|-\log\left|\bm{A}_n^{-1}\bm{R}_n\right|\right)
\end{multline} 

This implies that only $\mathcal{O}\left(n^3\right)$ matrix decompositions are required, which could represent huge savings compared to $\mathcal{O}\left(N^3\right)$ if the degree of replication is high. 

\begin{figure}[h]
    \centering
    {\includegraphics[width=0.95\linewidth]{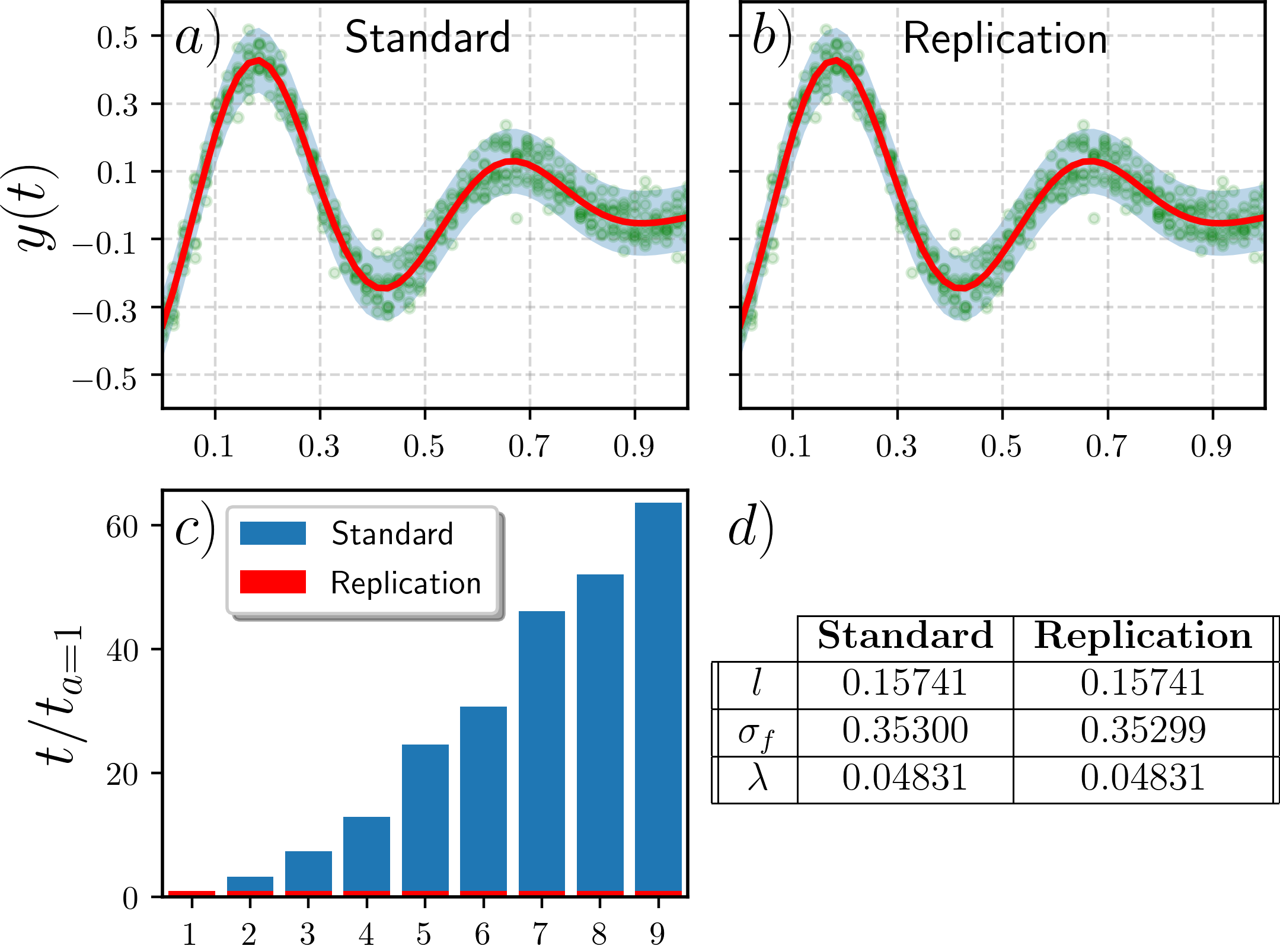}}
    \vspace{-3mm}
    \caption{(a) Learned policy with the standard GP formulation. The red solid line, blue shaded area and green dots refer respectively to the mean, the 95\% confidence interval and the training samples. (b) Same with the replication-based design. (c) Computational time for retrieving the policy with respect to $a=1$ for different $a$. (d) Inferred hyperparameters of the SE kernel, $l$ and $\sigma_f$, and noise variance $\lambda$.}
    \label{fig5}
\end{figure}

For illustrating the potential of exploiting the structure of replications in the design, consider a manipulation task with the following observational model
\begin{equation}
    y\left(t\right)=\sin\left(\pi\left[4t-0.25\right]\right)e^{-3t}\left(1+e^{-5x}\right)^{-1}+\epsilon
\end{equation} where $\epsilon\sim\mathcal{N}\left(0,\lambda=0.05\right)$. For the model, we assume an SE kernel with hyperparameters $l$ and $\sigma_f$, and constant noise level with variance $\lambda$, which we also consider a hyperparameter. Samples are taken at $n=50$ unique input locations, each having $a$ replicates. In Figure \ref{fig5}c, we can observe that for the standard formulation, the computational time for retrieving the policy with respect to $a=1$ increases dramatically with the number of replicates, whereas it remains constant with the proposed one. For the case $a=9$, where the time differs by a factor of 60, we can observe in figures \ref{fig5}a and \ref{fig5}b that the learned policy is identical. Also, the inferred hyperparameters (Figure \ref{fig5}d) since the method is exact.

\section{Illustrative Example: Learning to Write}
\label{experiment}

In this section, we illustrate and evaluate the main aspects of the presented GP-based LfD framework through the robot writing task. This skill suits perfectly the LfD context since it is difficult to script but can be intuitively demonstrated. The task has been explored for engaging robots in teaching activities. Building up on the learning by teaching paradigm, letting a child demonstrate the robot, not only does the child practice handwriting but, also positively reinforce their motivation \cite{Lemaignan2016}. First, we present the robot an experimental handwritten dataset, which includes trajectories for several letters indexed by real, integer and categorical task variables. Then, we consider the problem of learning a movement policy from the demonstrations, comparing the performance of different GP designs. Next, we assess the adaptability of the resulting policy by evaluating the modulation through via-points, and the interpolation and extrapolation capabilities. Finally, we study, in terms of computational time, the implications of considering the structure of replications in the formulation.

\subsection{Handwritten Letter Dataset}

\begin{figure}[h]
    \centering
    {\includegraphics[width=1.0\linewidth]{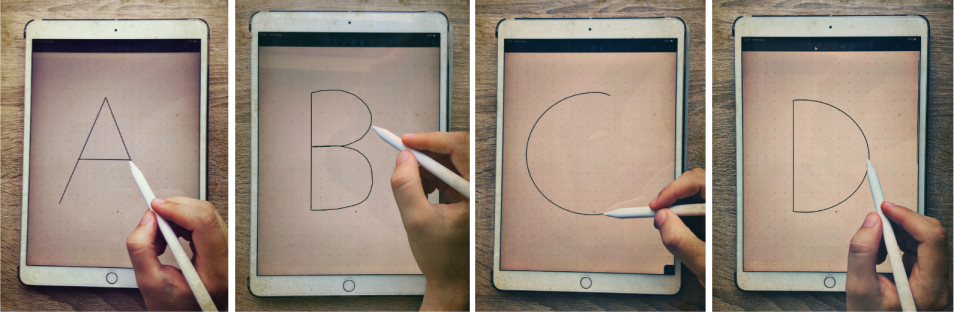}}
    %\vspace{-5mm}
    \caption{Demonstrations for the writing task are performed using a tablet.}
    \label{fig6}
\end{figure}

The demonstration dataset has been generated experimentally by handwriting different letters on a tablet using a standard note app (Figure \ref{fig6}). We extracted the data by first screen recording while writing, and then, processing the resulting videos with computer vision techniques. As output variables we take the $x$ and $y$ coordinates of the path that describes the handwriting motion. As input (or task) variables we take time $t$; the size of the letter $s$, which we consider defined by the height, measured as an integer $\text{height}=s\times8\text{mm}$; and finally, the letter corresponding to the motion $u$, e.g. 'A', 'B' ,'C', etc. The variables and the domain sampled in the dataset are summarized in Table \ref{tab2}.

\begin{table}[h]
\centering
\small
\begin{tabular}{||c|c|c|c||} 
 \hline
 \textbf{Variable} & \textbf{Symbol} & \textbf{Type} & \textbf{Domain} \\
 \hline\hline
 Time & $t$ & Input & $\left[0,\,1\right]$ \\
 \hline
 Size & $s$ & Input & $\left\{2,3,4,5,6\right\}$ \\
 \hline
 Letter & $u$ & Input & $\left\{\text{A},\text{B},\text{C},\text{D}\right\}$  \\
 \hline
 Horizontal coordinate & $x$ & Output & $\mathbb{R}$ \\
 \hline
 Vertical coordinate & $y$ & Output & $\mathbb{R}$  \\
 \hline
\end{tabular}
\caption{Handwritten letter dataset variables.}
\label{tab2}
\vspace{-5mm}
\end{table}

\begin{figure*}[t]
    \centering
    {\includegraphics[width=0.98\linewidth]{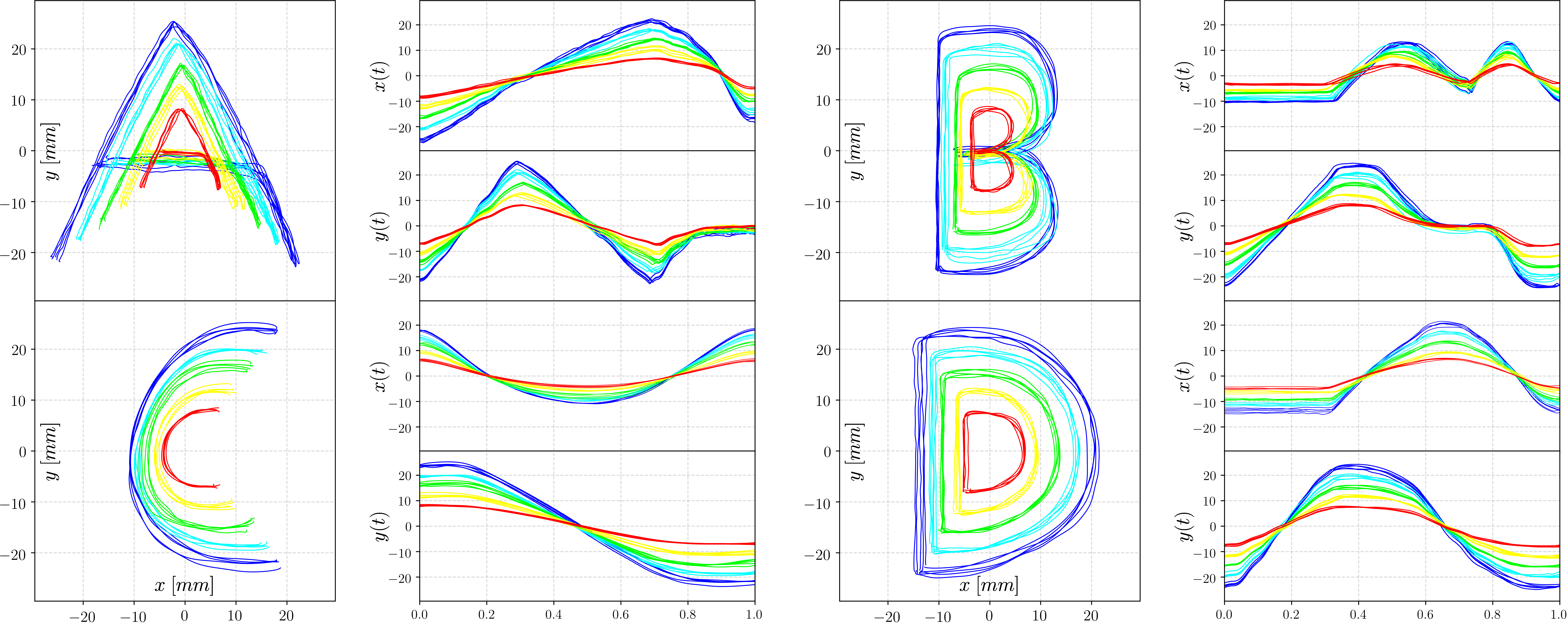}}
    \caption{Handwritten letter dataset. Trajectories are indexed by time $t$, size $s$ and letter $u$. 5 replicates are provided for each possible combination of the discrete variables $s$ and $u$, represented in the same color. To the right of each letter $x$-$y$ path, the corresponding $x(t)$ and $y(t)$ are shown.}
    \label{fig7}
    \vspace{-4mm}
\end{figure*}

The size of the dataset was guided by the discrete task variables, such that each of the $5\times 4=20$ possible discrete input locations has $5$ replicates. That is a total of $N=100$ different demonstrations. The dataset, after temporal alignment and scaling of trajectories, is shown in Figure \ref{fig7}. 

\subsection{Task Model Design}

We aim to learn a movement policy from the demonstrations that is capable of generating the motion required to write a given letter with a specified size. In the presented GP approach the modeling effort is focused on the covariance structure. The multi-dimensional kernel function is built as a composition, which can be either sum, product or ANOVA, of one-dimensional ones. For $k_{\mathbb{R}}$ we can use the SE or the Mat\'ern; for $k_{\mathbb{Z}}$ we propose the cosine kernel and, due to the nature of the size variable, a linear transformation $T()$ (Eq. \ref{eq-linear-warping}); finally, for $k_{\mathbb{K}}$ we take the CS structure. To select the best model, we split the dataset in 50\% for learning the task policy and the remaining 50\% for assessing its performance. The results obtained for different covariance structures in terms of the $R^2$ coefficient are shown in Table \ref{tab5}. Although very accurate predictions are obtained either with the product or ANOVA composition, the best score is achieved by the ANOVA+Mat\'ern model. In Figure \ref{fig9}a we can see that this policy effectively retrieves the motion required for writing a letter 'A' of size 4, also encoding the variability.

\begin{table}[h]
%\vspace{-3mm}
\centering
\small
\def\arraystretch{1.2}
\begin{tabular}{||c||c|c|c||} 
\hline
\diagbox[innerwidth=1.7cm,innerrightsep=-10pt,outerrightsep=10pt]{$\quad k_{\mathbb{R}}$}{\footnotesize{\textbf{Composition}}} & \textbf{Product} & \textbf{Sum} & \textbf{ANOVA}  \\ \hline\hline
\textbf{SE} & 0.87 & 0.32 & 0.93  \\ \hline
\textbf{Mat\'ern} & 0.89 & 0.35 & 0.94 \\ \hline
\end{tabular}
\caption{$R^2$ coefficient for different covariance structures.}
\vspace{-7mm}
\label{tab5}
\end{table}

%\begin{figure}[b]
%    \vspace{-6mm}
%    \centering
%    {\includegraphics[width=1.0\linewidth]{figures/Figure8.png}}
%    \vspace{-8mm}
%    \caption{Matrix $\bm{K}$ constructed with ANOVA composition and $k_\mathbb{R}\equiv$ Mat\'ern for $x$ (a) and $y$ (b) coordinates.}
%    \label{fig8}
%\end{figure}

\subsection{Adaptability of the Task Policy}

\setcounter{figure}{8}
\begin{figure}[b]
    \centering
    \vspace{-4mm}
    {\includegraphics[width=1.0\linewidth]{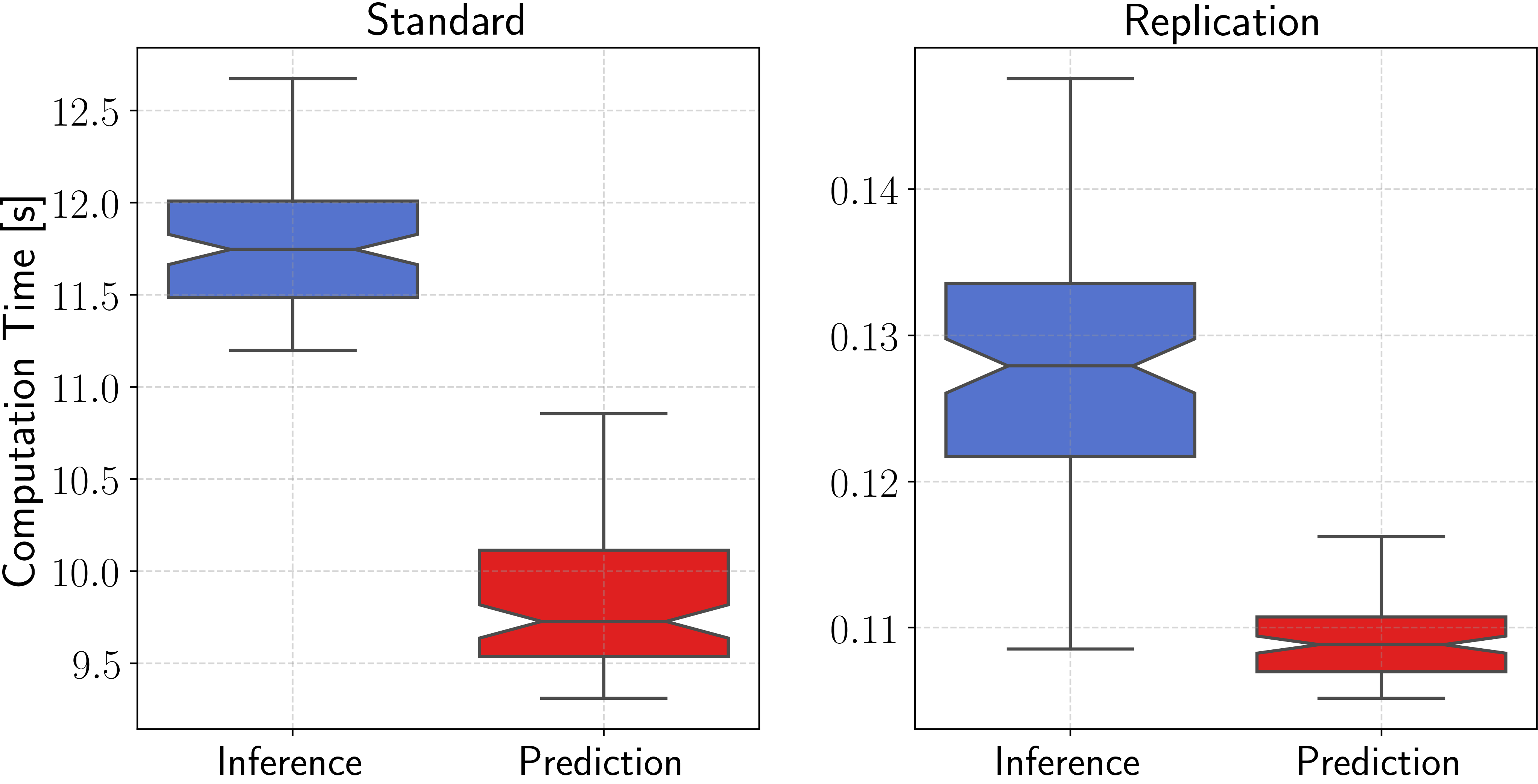}}
    \caption{Computational advantage of replication for evaluating the hyperparameter inference function, and calculating the predictive distribution.}
    \label{fig10}
\end{figure}

\setcounter{figure}{7}
\begin{figure*}[t]
    \centering
    {\includegraphics[width=1.0\linewidth]{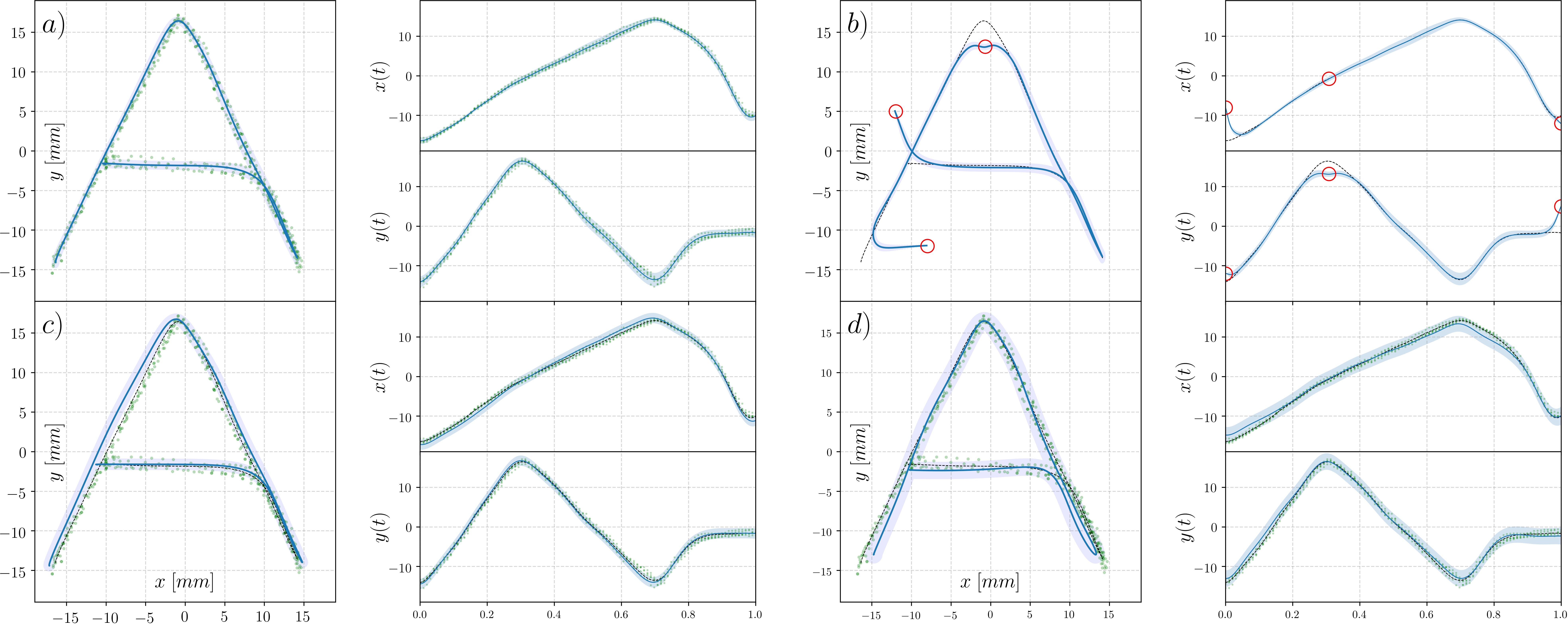}}
    \caption{\textbf{(a)} Motion retrieved by the task policy learned from the demonstrations for writing an 'A' of size 4. The solid blue line denotes the mean, the blue shaded area, the 95\% confidence interval, and the green dots, the training data. \textbf{(b)} Policy adaptation through via-points (red circles). For comparison, the mean of the previous case is represented as a black dashed line. \textbf{(c)} The model is built only with demonstrations of sizes 3 and 5, thus interpolation is required to generate the required motion. \textbf{(d)} In this case, only demonstrations of sizes 5 and 6 are used, thus extrapolation is performed.}
    \label{fig9}
    \vspace{-5mm}
\end{figure*}

Here we illustrate the generalization capabilities of the learned task policy. Adaptability can be achieved either through the specification of via-points, or relying on the GP model to generate the required motion given a new set of task parameters, for which demonstrations are not provided. We study different possibilities through the example shown in Figure \ref{fig9}. As depicted in Figure \ref{fig9}b, the motion in Figure \ref{fig9}a can be easily modulated to fulfill new specifications by conditioning the learned distribution to pass through new initial, final and/or intermediate via-points. Now consider that during the execution of the task, the robot encounters a new context, not sampled in the demonstration set. We consider two different cases, in Figure \ref{fig9}c we only use data of letters with sizes 3 and 5, whereas in Figure \ref{fig9}d we only consider sizes 5 and 6. Thus, for writing a letter 'A' of size 4 interpolation is required in the former case, and extrapolation in the latter. Since the new task variables are close to the demonstration region in both cases, we can see that the policies are capable of adapting effectively to the previously unseen scenarios.

\subsection{Computational Advantage of Replication}

Now we evaluate the potential benefit of exploiting the structure of replication during inference and prediction of the GP-based policy. From each of the 100 demonstrations we take 25 uniformly distributed timestamps for training. Thus, we have a total of $\mathcal{N}=2500$ points with $n=500$ unique input locations ($a=5$). In Figure \ref{fig10} we show a comparison of the computational time per evaluation of the inference function (inference), and for generating the task motion (prediction). We are able to perform the calculations 100 times faster without any approximation, achieving identical results.

\section{Conclusion}
\label{conclusion}

Learning from Demonstration (LfD) is arising as a promising paradigm that allows intuitively transferring manipulation skills to robots. A central problem is to design a movement policy such that the retrieved motion can automatically adapt to new situations encountered by the robot. Also, currently, as the complexity of the taught manipulation tasks increases, there is a growing need for learning algorithms capable of handling large demonstrations datasets.

In this paper, we present a Gaussian-Process-based LfD framework that allows an expressive and versatile encoding of the policy. We focus on generalization performance and computational efficiency. Adaptability is enhanced by incorporating task variables into the model, which can be either real, integer or categorical. On the other hand, the scalability of the learning algorithm is boosted by exploiting the structure of replications, which arise naturally in the LfD context. The proposed approach is illustrated and tested by teaching the robot how to write, achieving satisfactory performance in terms of policy design, adaptability and computational savings. An important future challenge will be to extend the framework towards learning more complex tasks such as cloth manipulation.

\bibliographystyle{IEEEtran}
\bibliography{References}

\end{document}